\documentclass[10pt,twocolumn,letterpaper]{article}

\PassOptionsToPackage{table}{xcolor}
\usepackage[pagenumbers]{cvpr}

\usepackage{multirow}
\usepackage{tabularx}
\usepackage{array}
\graphicspath{{./}{Figures/}}

\definecolor{cvprblue}{rgb}{0.21,0.49,0.74}
\usepackage[pagebackref,breaklinks,colorlinks,allcolors=cvprblue]{hyperref}

\def\confName{CVPR}
\def\confYear{2027}

\title{PathLoop: Self-Evolving Multi-Agent Reasoning for Evidence-Grounded Whole-Slide Image Analysis}
\author{
Jincheng~Li$^{1,2,3,*}$\quad Zhengyi~Liu$^{4,*}$\quad Zelin~Liu$^{5,6,*}$\quad Yihui~Zhan$^{7}$\quad Yifei~Sun$^{8}$\quad Xiangyu~Li$^{2,9}$\\
Minye~Shao$^{10}$\quad Lingyao~Kong$^{11}$\quad Tao~Hou$^{12}$\quad Xinheng~Lyu$^{6,13}$\quad Lanfeng~Zhong$^{2,14}$\\
Huamin~Wang$^{2,14}$\quad Ziquan~Zhu$^{15}$\quad Rongzhao~Zhang$^{2,\ddagger}$\quad Rui~Yan$^{1,3,\ddagger}$\quad S~Kevin~Zhou$^{1,3,\ddagger}$\\[5pt]
{\small $^{1}$ University of Science and Technology of China}\hspace{5pt}{\small $^{2}$ Shanghai Artificial Intelligence Laboratory}\hspace{5pt}{\small $^{3}$ MIRACLE Lab}\\
{\small $^{4}$ The Hong Kong University of Science and Technology (Guangzhou)}\hspace{5pt}{\small $^{5}$ Shanghai Jiaotong University}\\
{\small $^{6}$ Ant Group}\hspace{5pt}{\small $^{7}$ Nantong University}\hspace{5pt}{\small $^{8}$ Zhejiang University}\hspace{5pt}{\small $^{9}$ Tianjin University}\hspace{5pt}{\small $^{10}$ Durham University}\\
{\small $^{11}$ Jinan University}\hspace{5pt}{\small $^{12}$ Southeast University}\hspace{5pt}{\small $^{13}$ The University of Nottingham Ningbo China}\\
{\small $^{14}$ University of Electronic Science and Technology of China}\hspace{5pt}{\small $^{15}$ University of Leicester}\\[4pt]
{\small $^{*}$ These authors contributed equally to this work.\quad $^{\ddagger}$ Corresponding authors.}
}

\hypersetup{
    pdftitle={PathLoop: Self-Evolving Multi-Agent Reasoning for Evidence-Grounded Whole-Slide Image Analysis},
    pdfauthor={Jincheng Li, Zhengyi Liu, Zelin Liu, Yihui Zhan, Yifei Sun, Xiangyu Li, Minye Shao, Lingyao Kong, Tao Hou, Xinheng Lyu, Lanfeng Zhong, Huamin Wang, Ziquan Zhu, Rongzhao Zhang, Rui Yan, S Kevin Zhou}
}

\begin{document}

\maketitle

\begin{abstract}

Analyzing whole-slide images (WSIs) requires combining evidence across tissue regions and spatial scales. Multi-agent systems offer a promising approach to this task. However, existing systems make limited use of molecular evidence, often select visually prominent but less representative regions while overlooking global context, and struggle to reuse experience from previous cases. To address these challenges, we present PathLoop, a self-evolving multi-agent framework that integrates omics data for evidence-grounded WSI analysis. Specifically, PathLoop coordinates visual and omics interpreters across slide, patch, and pixel levels, using molecular signals to guide region selection and combining local tissue details with global slide features. A judgment agent then checks the consistency of the collected evidence against Pathology-Wiki, a knowledge base built from 48 entries across 24 organ systems, and triggers re-examination when the evidence is inconsistent. To reuse past experience, a four-level memory stores reasoning trajectories, practical tips, skills, and tools to guide future cases, forming a loop of evidence collection, review, and memory update. We evaluate PathLoop on four TCGA cohorts with biomarker prediction, survival prediction, and subtype classification. The results demonstrate the effectiveness of PathLoop across these tasks and support its use for evidence-grounded WSI analysis. 
\end{abstract}

\begin{figure}[!t]
\centering
\includegraphics[width=\columnwidth]{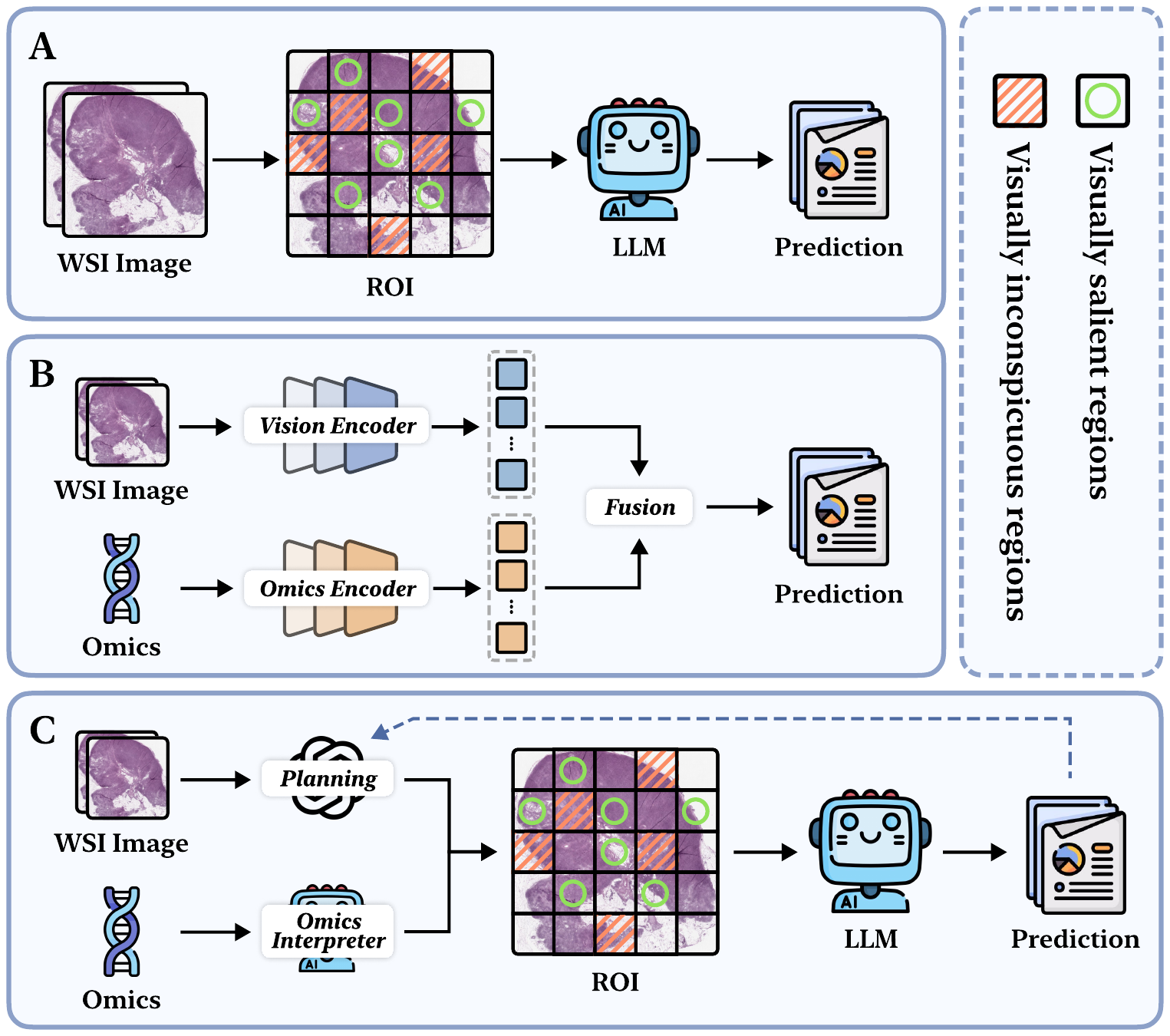}
\caption{Comparison between existing WSI analysis paradigms and PathLoop. (A) WSI-only methods rely primarily on visually salient ROIs and may miss diagnostically informative but visually inconspicuous regions. (B) Existing WSI–omics methods fuse separately encoded visual and molecular features, without using omics to guide ROI selection. (C) PathLoop leverages omics-informed planning to select task-relevant ROIs and introduces a feedback loop from prediction to planning for iterative refinement.}
\label{fig1}
\end{figure}

\section{Introduction}
Pathology is the gold standard for cancer diagnosis, and whole slide images (WSIs) show tissue specimens at gigapixel resolution and are critical for treatment decisions \cite{Li2025HighPrecisionMF, Li2026TwoStageCC}. However, owing to the complexity and fine-grained nature of pathological data, an ideal model must identify sparse diagnostic clues within extremely large images while navigating and interacting across multiple regions, posing substantially greater challenges than other medical imaging tasks \cite{Jiang2026UniCASAF}. Unlike the single-pass image-reading process commonly performed by radiologists, pathological diagnosis is inherently a long-horizon, interactive cognitive process. Collaborative multi-agent systems therefore constitute a natural paradigm for modeling this process and have recently attracted increasing attention \cite{Sun2023PathAsstAG, Sun2024PathGen16M1M}.

Existing pathology multi-agent systems largely emulate the workflow of pathologists: they first scan slides at low magnification to retrieve initial regions of interest (ROIs), progressively focus on these regions at higher magnifications, and conduct evidence retrieval under the guidance of specific diagnostic queries \cite{Lyu2025WSIAgentsAC, Sun2025CPathAgentAA, Wang2025PathologyCoTLV, Chen2025PathAgentTI, Wang2026QCAgentAA}. Their performance therefore depends on whether the selected regions contain evidence relevant to the target. This is difficult in heterogeneous tumors, where \textbf{visually prominent regions are not always the most informative}. Mainstream approaches  can be broadly divided into two categories: (i) weakly supervised multiple-instance learning (MIL) aggregates patches under slide-level labels and the attention is post-hoc and does not verify evidence sufficiency \cite{Ilse2018AttentionbasedDM, Shao2021TransMILTB}; (ii) although VLM-based WSIs readers make the retrieval process explicit, they rely primarily on prompt-level guidance and lack molecularly informed retrieval priors, making them prone to selecting visually salient regions that are irrelevant to the diagnostic endpoint \cite{Liang2024WSILLaVAAM, Tang2023MedAgentsLL}. Crucially, a related challenge is to retain slide-level context when interpreting local tissue details, since evidence from an isolated region may not reflect patterns across the full slide. \cite{Naeem2024PathRAGKK}.

Another limitation is that current pathology multi-agent systems \textbf{cannot effectively leverage omic information for evidence-grounded diagnosis}. Omic data and histopathological morphology provide complementary perspectives on tumor biology and evolution: morphological insights reveal tissue architecture and microenvironment, while molecular features from omics elucidate the underlying biological mechanisms \cite{Yan2025PathwayAwareMT, Xu2024AMK}. However, most existing  systems operate exclusively on images. The few multimodal WSI-omics fusion methods \cite{Huang2025SurvAgentHC, Jaume2023ModelingDM, Chen2021MultimodalCT} treat omics merely as a parallel feature stream or supplementary text. Omics neither guides region retrieval nor verifies the completeness of the retrieved evidence, resulting in final predictions with limited biological relevance and insufficient accuracy.

A third challenge is that pathology multi-agent systems still \textbf{analyze each test case independently, overlooking valuable diagnostic and prognostic experience accumulated from previous cases}. From a cognitive perspective, clinical expertise develops through cumulative cross-case experience: prior cases are recalled, compared, and progressively abstracted into reusable patterns that guide future decisions \cite{hu2025landscape}. By integrating semantic knowledge, episodic memory, and procedural knowledge, this cognitive process enables clinicians to adapt to novel scenarios that are not fully covered by training data or textbooks. Therefore, current agents require a new evolutionary paradigm that integrates experiential learning with explicit reasoning trajectories to enable effective WSIs analysis.

To address these challenges, we propose PathLoop, a self-evolving,
evidence-grounded, omics-enhanced multi-agent framework for WSI analysis.
The framework comprises four collaborative agents: Planning, Execution,
Judgment, and Persist, which decompose long-horizon WSI analysis into
specialized cognitive stages to mitigate cognitive overload.
The Planning Agent interprets the task, identifies the required data
modalities, retrieves relevant prior experience, and assigns subtasks
to specialist modules.
The Execution Agent coordinates visual and omics experts across slide,
patch, and pixel levels, using molecular signals to guide region
selection and integrating slide-level context with fine-grained visual
evidence.
The Judgment Agent evaluates the consistency of the collected evidence
against Pathology-Wiki and triggers re-examination when inconsistencies
arise, thereby enabling evidence-grounded self-correction.
Finally, the Persist Agent maintains a four-level memory of reasoning
trajectories, practical tips, skills, and tools
\cite{Zhang2026OPDEvolverCH, Feng2026MedExpMemAE, Cao2025RememberMR}.
It updates this memory using task outcomes and post hoc diagnostic
ground truth, enabling subsequent cases to benefit from prior experience
without parameter updates.
Together, these agents form a closed loop of evidence collection,
verification, and experience reuse.

Our main contributions are summarized as follows:
\begin{itemize}
  \item We propose PathLoop, a self-evolving, evidence-grounded, omics-enhanced multi-agent framework for computational pathology, which decomposes long-horizon WSIs diagnosis into procedural cognitive steps via Planning, Execution, Judgment, and Persist Agents, alleviating cognitive overload. To our knowledge, it is the first pathology agent that uses omics to guide region selection and accumulates reusable experience across cases.

  \item We design an omics-guided retrieval mechanism that uses molecular signals to select diagnostically relevant regions, and a four-level memory that improves future cases from past experience without parameter updates.

  \item We curate and publicly release Pathology-Wiki, constructed from 48 entries across 24 organ systems. On four TCGA cohorts covering biomarker prediction, survival prediction, and subtype classification, PathLoop consistently outperforms pathology MLLMs, pathology agents, and WSI--omics fusion methods.
\end{itemize}

\begin{figure*}[!t]
\centering
\includegraphics[width=\textwidth]{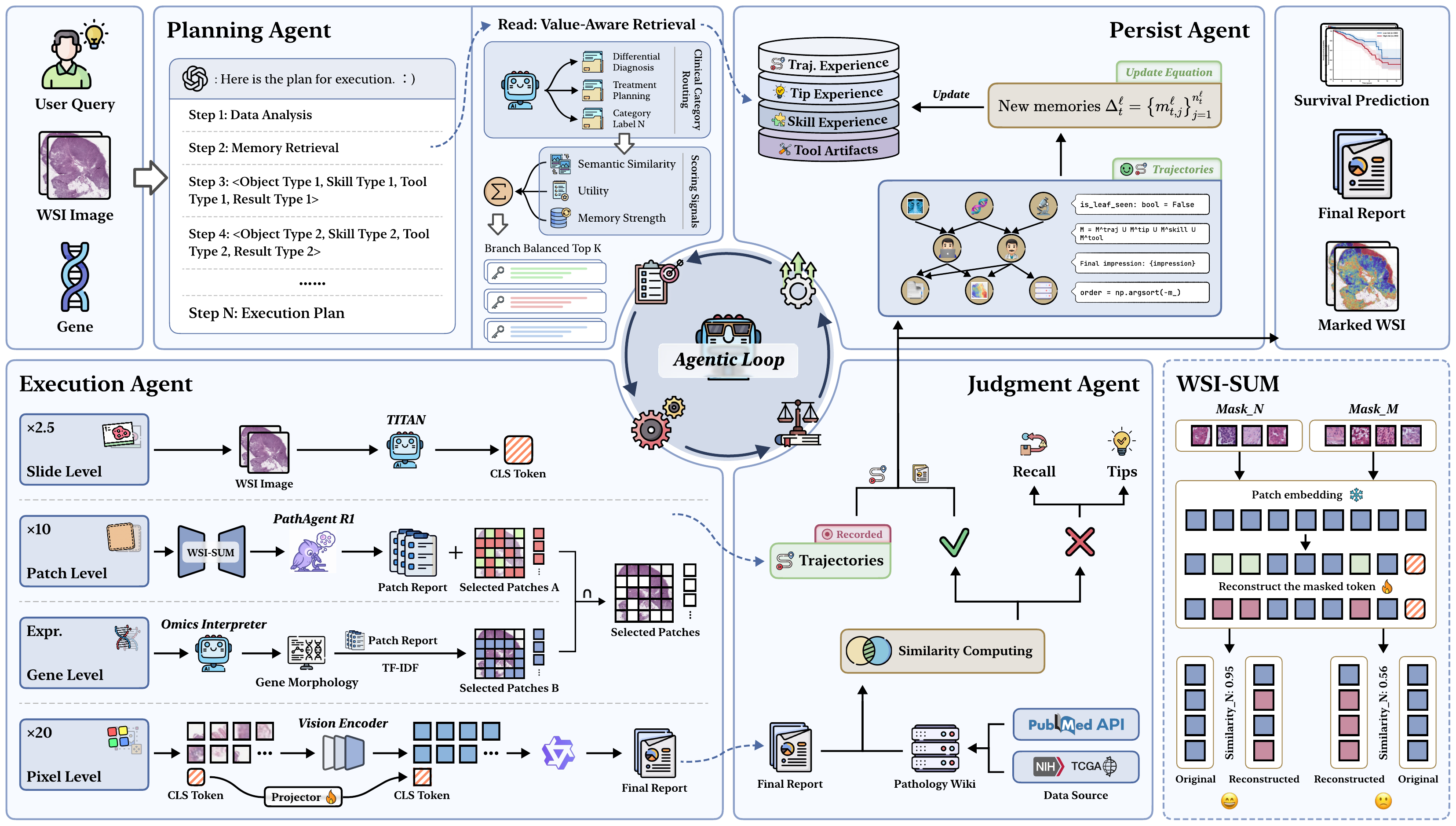}
\caption{The PathLoop framework comprises four alternating phases. The Planning Agent analyzes the input and formulates a plan by drawing on historical experience.
The Execution Agent performs slide-, patch-, and pixel-level processing and integrates omics data.
The Judgment Agent conducts a consistency review against Pathology-Wiki.
Finally, the Persist Agent builds a four-level memory from past trajectories, which is retrieved to guide reasoning in future cases.
}

\label{fig2}
\end{figure*}

\section{Related Work}
\subsection{WSIs Analysis Techniques}
With the development of LLMs, WSIs analysis has evolved from MIL toward a new paradigm based on MLLMs. Recent pathology-specific MLLMs have demonstrated substantial potential for WSIs analysis. CONCH \cite{Lu2024AVF}, MUSK \cite{Xiang2025AVF}, and PLIP \cite{Huang2023AVF} achieve strong performance in patch retrieval through vision--language contrastive learning, while Quilt-LLaVA \cite{Seyfioglu2023QuiltLLaVAVI} and Hepato-LLaVA \cite{Yang2026HepatoLLaVAAE} further incorporate visual evidence into open-ended descriptions, exhibiting enhanced reasoning capabilities. However, constrained by limited context windows, these models only local patches. To overcome this limitation, SlideChat \cite{Chen2024SlideChatAL}, WSI-LLaVA \cite{Liang2024WSILLaVAAM}, and WSI-SUM \cite{Wang2026WSISumWS} attempt to analyze complete WSIs through patch aggregation, yet they remain insufficient in providing well-grounded visual reasoning and comprehensive diagnostic planning. More fundamentally, existing MLLMs typically process pathological inputs through a fixed feed-forward pipeline, which is inherently inconsistent with the dynamic, interactive, and multiscale diagnostic workflow of expert pathologists.

\subsection{Pathology Multi-Agent Systems}
Gigapixel WSIs analysis is inherently complex, requiring cross-scale navigation and cross-region interaction. Collaborative multi-agent systems thus offer a promising paradigm for such long-horizon diagnostic processes. For instance, SlideSeek ~\cite{Chen2025EvidencebasedDR} coordinates a supervisor, multiple parallel explorers, and a reporter for autonomous whole-slide analysis. CPathAgent~\cite{Sun2025CPathAgentAA} simulates pathologists' multiscale reasoning through a three-stage navigation--reasoning pipeline. PathAgent~\cite{Chen2025PathAgentTI} constructs explicit thought chains from sequential observations and decisions for interpretable analysis. SurvAgent~\cite{Huang2025SurvAgentHC} combines LMScreen, CoSMining, and ConfMining across scales for survival prediction, forming an effective multi-agent framework.

PathLoop addresses these challenges by combining omics-guided region retrieval, cross-scale visual analysis, evidence checking against Pathology-Wiki, and memory updates across cases. Rather than treating region selection, evidence review, and experience reuse as separate steps, PathLoop brings them into a single multi-agent workflow for WSI analysis.

\section{Method}
PathLoop decomposes WSI analysis into four stages and each handled by a dedicated agent (Figure~\ref{fig2}).
We describe the four agents in turn.

\subsection{Planning Agent} 
Given the input WSI image $I$, gene data $G$, and user query $Q$, the Planning Agent based on GPT-5.5 \cite{openai2026gpt55} first reformulates the query into a structured prompt $P$, and dynamically dispatches relevant expert subsets by mapping $(I,P)$ and $(G,P)$. The system maintains a four-level memory bank, storing $\mathcal{L}=\{\text{traj}, \text{tip}, \text{skill}, \text{tool}\}$, where each $\mathcal{M}^{l}_{t}$ stores type-$l$ memories: $\mathcal{M}_{t}=\mathcal{M}^{\text{traj}}_{t}\cup\mathcal{M}^{\text{tip}}_{t}\cup\mathcal{M}^{\text{skill}}_{t}\cup\mathcal{M}^{\text{tool}}_{t}$, and the specific implementation will be described in the Persist Agent section. We form the query $z_t=[P_t:I_t:G_t]$, and retrieve the top-$K$ candidates from each memory level through embedding similarity:
\begin{equation}
\mathcal{C}^{l}_{t}=\mathrm{TopK}_{m\in\mathcal{M}^{l}_{t}}\mathrm{sim}(\phi(z_t),\phi(m)),\quad C_t=\bigcup_{l\in\mathcal{L}}\mathcal{C}^{l}_{t}.
\end{equation}
where $\phi(\cdot)$ is the embedding encoder, $\mathrm{sim}(\cdot)$ denotes cosine similarity, and $\mathrm{TopK}$ returns the $K$ memories with the highest scores. The selector then compresses $C_t$ into the memory context presented to the agent:
\begin{equation}
S^{l}_{t}=\mathrm{Sel}^{l}_{\theta}(x_t,C^{l}_{t}),\quad S_t=\bigcup_{l\in\mathcal{L}}S^{l}_{t},\quad c_t=\mathrm{Fmt}(S_t).
\end{equation}

Based on $z_t$ and $c^{l}_{t}$, the VLM generates a disease-specific diagnostic plan containing multiple reasoning steps $\mathcal{P}=\{P_1,P_2,\cdots,P_n\}$. Each step $P_i\in\mathcal{P}$ is represented as:
\begin{equation}
P_i:r_i=\langle k_{r_i},v_{r_i}\rangle=\langle k_{r_i},t(v_{o_i})\rangle,\quad t\in\mathcal{T}.
\end{equation}

The processed object $o_i=(k_o,o_o)$ and the result $r_i=(k_r,O_r)$ are represented as key-value pairs, where the key ($k_o$ or $k_{rs}$) specifies the data attribute, and the value ($U_o$ or $U_{rs}$) contains the actual data. The specific implementation is described in \cite{Wang2025MedAgentProTE}.

\subsection{Execution Agent}
\paragraph{Slide Level:} 
Given a WSI $I$, we downsample it to $I_{2.5}$ and compute a 768-dimensional slide-level embedding $\mathbf{s}$ using the TITAN \cite{Ding2025AMW} encoder, summarizing the global scan.

\paragraph{Patch Level:} 
We use a sliding window to partition WSIs into $10\times$ patches and encode them with CONCHv1.5~\cite{Lu2024AVF} as $X = \mathrm{PatchEncoder}(P)$, where $P \in \mathbb{R}^{N\times 3\times 2048\times 2048}$ and $X \in \mathbb{R}^{N\times C}$.
Following~\cite{Wang2026WSISumWS}, we identify redundant patch embeddings by their reconstructability from unmasked ones. We partition $X$ into $K$ clusters via K-means, obtaining $X_k = M_k X$. Within each cluster, we randomly mask a subset $M \subset X_k$ and feed the unmasked tokens $X_k \setminus M$ with positional encodings $P_v$ to an encoder: $Z = \mathrm{Encoder}(X_k \setminus M + P_v)$. A lightweight decoder reconstructs the masked embeddings from $Z$, learnable mask tokens $m$, and their positional encodings $P_m$: $\hat{X} = \mathrm{Decoder}(Z, m, P_m)$, with $\hat{X}=\{e_i\}$ maintaining the original order. Path-R1 then analyzes patch sequence representativeness and generates preliminary reports $\{R_i\}_{i=1}^{n}$, yielding selected lesion ROI records $P_{select}^{v}$.

Following \cite{Li2026HEROHE}, given DNA methylation vector $\mathbf{x}_{dna}\in\mathbb{R}^{F_d}$ and miRNA vector $\mathbf{x}_{mirna}$, we standardize them and project to $P\!=\!50$ pathways using predefined binary masks $\mathbf{M}_{dna}\!\in\!\mathbb{R}^{P\times F_d}$ (MSigDB Hallmark) and $\mathbf{M}_{mirna}\!\in\!\mathbb{R}^{P\times F_m}$ (miRTarBase). The pathway score is $s_p = s^p_{dna} + s^p_{mirna}$, with $s^p_{dna} = -\,\frac{\sum_i M^p_{dna,i}\,z(x^i_{dna})}{\sum_i M^p_{dna,i} + \epsilon}$ and $s^p_{mirna}$ analogous. The negative sign indicates downregulation, and the denominator normalizes by pathway member count.

\begin{figure*}[!t]
\centering
\includegraphics[width=\textwidth]{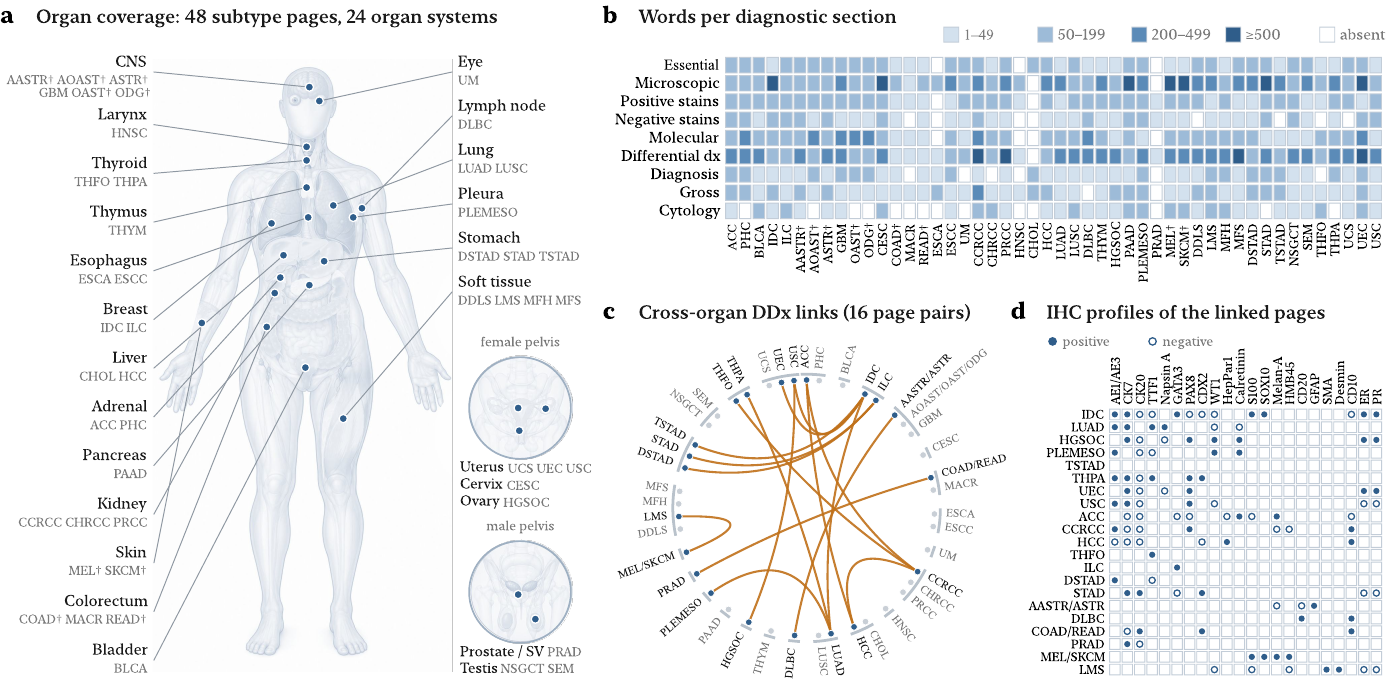}
\caption{Overview of Pathology-Wiki. (a) Organ coverage of the knowledge base, spanning 24 organ systems and 48 subtype codes. Codes marked with $\dagger$ share content with another code, resulting in 43 distinct pages. (b) Word counts for major diagnostic sections in each page, showing the completeness and depth of the curated content. (c) Cross-organ differential-diagnosis DDx links among 43 pages, highlighting 16 page pairs that reference entities from different organs. (d) IHC profiles of the linked pages, summarizing positive and negative markers that support differential diagnosis.}
\label{fig3}
\end{figure*}

We define a sparse matrix $\mathbf{W}\!\in\!\mathbb{R}^{K\times P}$ ($K\!=\!16$) with $\ell_1$ row normalization, whose rows correspond to morphological axes (e.g., mitotic activity, necrosis, stromal fibrosis) and columns to pathways. Entries are set based on WHO terminology and biological associations, without optimization against endpoint labels. The intention vector is computed as
\begin{equation}
\mathbf{m}=\ell_2\text{-normalize}\left(\operatorname{ReLU}(\mathbf{W}\mathbf{s})\right)\in\mathbb{R}^{K},
\end{equation}
where $\mathbf{s}=[s_1,\dots,s_P]^\top$. ReLU suppresses negative contributions, and $\ell_2$ enables cosine similarity comparisons.

From $\mathbf{m}$, a committee of three role-conditioned LLMs generates a keyword set $Q$ (15--40 terms) to guide endpoint-aware retrieval.
We extract $256\!\times\!256$\,px patches at $10\times$, encode them into $1024$-d $\ell_2$-normalized embeddings with UNI, and apply MiniBatchKMeans ($k\!=\!300$) to select the patch nearest each centroid as a representative. Each representative is captioned by Qwen3-VL-8B at $10\times$, producing structured captions $\text{cap}^{10}_i$ covering tumor architecture, stroma and necrosis. A TF-IDF vectorizer $\phi$ is fit on the $300$ captions per slide. We rank patches by $\text{score}^{10}(p_i)=\cos(\phi(Q),\phi(\text{cap}^{10}_i))$, select the Top-64 to form an $8\!\times\!8$ mosaic for VLM inference in a single forward pass. Subpatches $\mathcal{P}_{20}=\bigcup_k \mathcal{P}^{20}_k$ are extracted via cross-modal similarity filtering and magnified to $20\times$ for fine-grained review.

\paragraph{Pixel Level:} 
For ROIs, we magnify them to $\times20$ for fine-grained assessment. To integrate the global CLS token, we train a lightweight projector: from the slide-level embedding $\mathbf{s}$, a two-layer MLP projects it to a global token $\mathbf{g} = \operatorname{MLP}(\mathbf{s})$ in Qwen3-VL-8B's token space. We inject this descriptor by reserving a special marker token before the ROI images and overwriting its embedding with $\mathbf{g}$ via a forward hook, leaving sequence lengths, M-RoPE positions, and the attention mask unchanged. The CLS token then co-attends with the fine-grained ROI patch tokens, supplying whole-slide context to the zoomed reading. This stage produces the final report.

\subsection{Judgment Agent}

The Judgment Agent verifies diagnoses from the Execution Agent by consulting a structured Pathology-Wiki rather than relying on embedding-based retrieval. The wiki is constructed from TCGA reports and PubMed, and contains 48 subtype codes across 24 organ systems and 27 tumor types; after merging codes that share the same content, this corresponds to 43 distinct JSON pages. Each page follows a standardized schema with fields such as Essential Features, Microscopic Description, Immunohistochemistry, and Differential Diagnosis. Figure~\ref{fig3} summarizes the scope, content structure, and diagnostic utility of Pathology-Wiki.

Given a diagnostic text $d$, the best-matching subtype $c$ is resolved via lexical routing:
\begin{equation}
\operatorname{resolve}(d)=\arg\max_{c\in\mathcal{W}} s(d,c),
\end{equation}
where the matching score $s(d,c)$ is a weighted sum: 6 points for a name/alias match, 2 for an organ match, and 1 per overlapping term.
The system retrieves the diagnostic card $\operatorname{card}(c)$ and its differential-diagnosis cards, then prompts the language model with the concatenation of the original description, matched card, and differential cards.
The model outputs a JSON verdict containing \textit{pass/revise}, \textit{supported}, \textit{missing}, \textit{preferred}, \textit{diff}, \textit{corrected}, and \textit{citations}; all claims cite the knowledge base verbatim.
When inference is unavailable, a deterministic fallback is applied: if the top-matching score exceeds $\tau_{\mathrm{route}}$ and no same-organ competitor lies within $\Delta$ points, the verdict is \textit{pass}; otherwise \textit{revise}.
If revised, the wiki-based feedback is returned to the Execution Agent, triggering a $20\times$ intensified re-examination and diagnosis regeneration.
This forms a Judgment-Execution closed loop, capped at $L$ iterations, which leverages structured knowledge to correct overconfident predictions and misclassifications.

\begin{table*}[!t]
\centering
\small
\setlength{\aboverulesep}{0pt}
\setlength{\belowrulesep}{0pt}
\renewcommand{\arraystretch}{1.2}
\newcolumntype{Y}{>{\centering\arraybackslash}X}
\caption{Comparison with SOTA methods on multi-task prediction on TCGA-BRCA.
AUC for ER/PR/HER2, macro AUC for Subtype, and C-index for Risk.
Modality denotes the input data: O for omics, W for WSI, and O+W for both.
The best and second-best results are highlighted in
\textbf{bold} and \protect\underline{underlined}, respectively.}
\label{tab:comparison}
\begin{tabularx}{\textwidth}{lcYYYYY}
\toprule
\textbf{Method}
& \textbf{Modality}
& \textbf{ER (AUC)}
& \textbf{PR (AUC)}
& \textbf{HER2 (AUC)}
& \textbf{Subtype (AUC)}
& \textbf{Risk (C-index)} \\
\midrule
\rowcolor{gray!15}
\multicolumn{7}{l}{$\blacktriangledown$~\textit{Conventional Methods}} \\
\midrule
SNN
& O
& 0.952$\pm$0.021
& 0.885$\pm$0.035
& 0.725$\pm$0.045
& 0.958$\pm$0.015
& 0.620$\pm$0.045 \\
SNNTrans
& O
& 0.958$\pm$0.018
& 0.912$\pm$0.028
& 0.765$\pm$0.038
& 0.970$\pm$0.012
& 0.645$\pm$0.038 \\
ABMIL
& W
& 0.885$\pm$0.032
& 0.755$\pm$0.045
& 0.655$\pm$0.052
& 0.955$\pm$0.022
& 0.590$\pm$0.052 \\
TransMIL
& W
& 0.901$\pm$0.028
& 0.763$\pm$0.041
& 0.687$\pm$0.048
& 0.972$\pm$0.015
& 0.615$\pm$0.048 \\
MCAT
& O+W
& \textbf{0.975$\pm$0.022}
& 0.935$\pm$0.031
& 0.768$\pm$0.042
& 0.975$\pm$0.018
& 0.665$\pm$0.042 \\
MOTCat
& O+W
& 0.962$\pm$0.019
& 0.948$\pm$0.025
& 0.795$\pm$0.036
& \underline{0.980$\pm$0.014}
& 0.685$\pm$0.035 \\
SurvPath
& O+W
& 0.941$\pm$0.015
& \underline{0.957$\pm$0.022}
& \underline{0.815$\pm$0.032}
& 0.975$\pm$0.012
& \underline{0.705$\pm$0.030} \\
\midrule
\rowcolor{gray!15}
\multicolumn{7}{l}{$\blacktriangledown$~\textit{Pathology Foundation Models}} \\
\midrule
CONCH
& W
& 0.845$\pm$0.004
& 0.758$\pm$0.007
& 0.574$\pm$0.014
& 0.772$\pm$0.005
& 0.612$\pm$0.011 \\
TITAN
& W
& 0.857$\pm$0.009
& 0.764$\pm$0.009
& 0.633$\pm$0.016
& 0.814$\pm$0.004
& 0.642$\pm$0.012 \\
QuiltNet
& W
& 0.765$\pm$0.012
& 0.674$\pm$0.013
& 0.611$\pm$0.025
& 0.714$\pm$0.010
& 0.560$\pm$0.015 \\
PLIP
& W
& 0.790$\pm$0.011
& 0.713$\pm$0.005
& 0.598$\pm$0.015
& 0.727$\pm$0.006
& 0.556$\pm$0.007 \\
\midrule
\rowcolor{gray!15}
\multicolumn{7}{l}{$\blacktriangledown$~\textit{Proprietary MLLMs}} \\
\midrule
GPT-5.5
& O+W
& 0.945$\pm$0.035
& 0.880$\pm$0.045
& 0.710$\pm$0.055
& 0.950$\pm$0.025
& 0.610$\pm$0.065 \\
Gemini-2.5Pro
& O+W
& 0.932$\pm$0.023
& 0.871$\pm$0.054
& 0.791$\pm$0.015
& 0.948$\pm$0.029
& 0.649$\pm$0.065 \\
\midrule
\rowcolor{gray!15}
\multicolumn{7}{l}{$\blacktriangledown$~\textit{Medical Agents}} \\
\midrule
MDAgent
& O+W
& 0.962$\pm$0.025
& 0.915$\pm$0.038
& 0.755$\pm$0.048
& 0.972$\pm$0.018
& 0.640$\pm$0.050 \\
\textbf{PathLoop (Ours)}
& O+W
& \underline{0.964$\pm$0.004}
& \textbf{0.958$\pm$0.016}
& \textbf{0.854$\pm$0.007}
& \textbf{0.983$\pm$0.022}
& \textbf{0.735$\pm$0.005} \\
\bottomrule
\end{tabularx}
\end{table*}

\subsection{Persist Agent}
The agent maintains a four-level memory of trajectories, tips, skills, and tools.
Trajectories form a slide-level tree; the optimal path maximizes leaf value:
\begin{equation}
\operatorname{best\_path}=\arg\max_{\rho}V(\operatorname{leaf}(\rho)), V(n) = \max_{\ell\preceq n}o(\ell).
\end{equation}

Memory operations consist of \textit{SELECT}, \textit{WRITE}, and \textit{MANAGE} stages.
In \textit{SELECT}, coarse retrieval first applies global embeddings, then retrieves top-$K$ by cosine similarity over flattened memory; matched memories are injected \textit{for reference only} to avoid contamination.
After execution, \textit{WRITE} distills the current decision process into the memory structure.
Every $Q$ runs, \textit{MANAGE} scores each memory item by $V(m)=\alpha_{\ell}\cdot\gamma(m)\cdot\hat{A}(m)$, merges near-duplicates and removes obsolete content for compactness.

The outcome signal is $o = \cos(E_{\mathrm{text}}(d), E_{\mathrm{text}}(\mathrm{GT}))$, with a consistency fallback when ground truth is absent.
Credit assignment uses a controlled-candidate advantage $\hat{A}(m)=\operatorname{mean}(o \mid m \text{ retrieved})-\operatorname{mean}(o \mid m \text{ not retrieved})$, which yields near-zero credit for always-retrieved memories, preventing erroneous assignment.
Only memories that improve outcome are reinforced, creating a cross-slide knowledge loop without parameter updates.

\section{Experiments}
\subsection{Experimental Dataset and Setup}
We evaluated PathLoop on four TCGA cohorts: BRCA, LUAD, BLCA, UCEC. We evaluate six tasks: ER, PR, HER2, subtype classification, survival risk prediction, and report generation accuracy. Report generation accuracy is assessed using CLIP-cos, BLEU-1, BLEU-4, METEOR, and ROUGE-L.
Labels are derived from TCGA clinical annotations; samples with missing or uncertain receptor status are excluded per task. 
Baselines include omics-only methods (SNN \cite{Klambauer2017SelfNormalizingNN} and SNNTrans \cite{Zhou2024CohortIndividualCL}), WSI-only MIL methods (ABMIL \cite{Ilse2018AttentionbasedDM} and TransMIL \cite{Shao2021TransMILTB}), multimodal fusion methods (MCAT \cite{Chen2021MultimodalCT}, MOTCat \cite{Xu2023MultimodalOT}, and SurvPath \cite{Jaume2023ModelingDM}), pathology VLM/agent methods (MDAgent \cite{Kim2024MDAgentsAA} and WSI-Agent \cite{Lyu2025WSIAgentsAC}), and general-purpose commercial models (GPT-5.5, Gemini-2.5Pro).  The projector is trained on an A100 GPU.

\begin{table}[!t]
\centering
\caption{Ablation study of the projector module and comparative experiments on language generation performance.}
\label{tab:projector_ablation}
\scriptsize
\renewcommand{\arraystretch}{1.12}
\setlength{\tabcolsep}{2pt}

\resizebox{\columnwidth}{!}{
\begin{tabular}{lccccc}
\toprule
Method
& \multicolumn{1}{c}{Alignment}
& \multicolumn{4}{c}{NLP Metrics} \\
\cmidrule(lr){2-2}
\cmidrule(lr){3-6}
& CLIP-cos 
& BLEU-1 
& BLEU-4 
& METEOR 
& ROUGE-L \\
\midrule
\textbf{w/ Projector}
& \textbf{0.66}
& \textbf{0.53}
& \textbf{0.14}
& \textbf{0.17}
& \textbf{0.19} \\
w/o Projector
& 0.53
& 0.43
& 0.02
& 0.04
& 0.06 \\
\midrule
Quilt-LLaVA
& 0.485
& 0.47
& 0.03
& 0.07
& 0.06 \\
HistGen
& 0.51
& 0.08
& 0.01
& 0.10
& 0.10 \\
MDAgents
& 0.46
& 0.01
& 0.00
& 0.05
& 0.06 \\
MedAgents
& 0.53
& 0.15
& 0.03
& 0.11
& 0.08 \\
WSI-Agent
& 0.58
& 0.47
& 0.09
& 0.11
& 0.14 \\
\bottomrule
\end{tabular}
}
\end{table}

\subsection{Comparisons with Base Models}
\paragraph{Multi-task prediction and report generation (Table~\ref{tab:comparison}).}
For multi-task prediction, PathLoop attains the best PR (0.958), HER2 (0.854), Subtype (0.983), and Risk (0.735) among all compared methods. Relative to the previous strongest baseline SurvPath, it improves HER2 by 0.039 (from 0.815 to 0.854) and Risk by 0.030 (from 0.705 to 0.735), and further surpasses SurvPath on Subtype by 0.008 (0.983 vs.\ 0.975). For ER, MCAT achieves the highest AUC (0.975), while PathLoop obtains 0.964, ranking second and outperforming every other baseline.

In the report generation and alignment experiment (Table~\ref{tab:projector_ablation}), adding the projector improves the CLIP-cosine alignment score from 0.53 to 0.66.
The gains in NLP metrics are also substantial: BLEU-1 rises from 0.43 to 0.53, BLEU-4 from 0.02 to 0.14, METEOR from 0.04 to 0.17, and ROUGE-L from 0.06 to 0.19.
PathLoop further outperforms all existing report generation methods on every metric; relative to the strongest baseline WSI-Agent, it improves CLIP-cos by 0.08 (0.66 vs.\ 0.58), BLEU-4 by 0.05 (0.14 vs.\ 0.09), METEOR by 0.06 (0.17 vs.\ 0.11), and ROUGE-L by 0.05 (0.19 vs.\ 0.14).
These results indicate that the projector effectively bridges visual and textual modalities for pathology report generation.

To verify that PathLoop generalizes beyond a single cohort, we additionally evaluate survival risk prediction on BLCA, LUAD, and UCEC (Table~\ref{tab:risk}).
PathLoop achieves the best mean C-index (0.695) across the four cohorts, ranking first on BRCA and UCEC and second on LUAD, indicating that evidence-grounded reasoning transfers across cancer types rather than overfitting to the primary cohort.

\begin{table}[!t]
\centering
\caption{Comparison of survival risk prediction (C-index) on four TCGA cohorts.
The best and second-best results are highlighted in
\textbf{bold} and \protect\underline{underlined}, respectively.}
\label{tab:risk}
\scriptsize
\renewcommand{\arraystretch}{1.12}
\setlength{\tabcolsep}{2pt}
\resizebox{\columnwidth}{!}{
\begin{tabular}{lccccc}
\toprule
Method
& BLCA
& LUAD
& BRCA
& UCEC
& Mean \\
\midrule
SNN
& 0.541$\pm$0.016
& 0.539$\pm$0.069
& 0.620$\pm$0.045
& 0.493$\pm$0.069
& 0.548 \\
SNNTrans
& 0.646$\pm$0.043
& 0.634$\pm$0.049
& 0.645$\pm$0.038
& 0.632$\pm$0.032
& 0.639 \\
TransMIL
& 0.576$\pm$0.065
& 0.615$\pm$0.034
& 0.615$\pm$0.048
& 0.549$\pm$0.044
& 0.589 \\
E2E-CONCH
& 0.653$\pm$0.024
& \textbf{0.705$\pm$0.029}
& 0.660$\pm$0.014
& 0.614$\pm$0.076
& 0.658 \\
MCAT
& 0.645$\pm$0.053
& 0.636$\pm$0.043
& 0.665$\pm$0.042
& 0.634$\pm$0.018
& 0.645 \\
MOTCat
& \underline{0.674$\pm$0.024}
& 0.674$\pm$0.036
& 0.685$\pm$0.035
& 0.667$\pm$0.051
& 0.675 \\
DPSurv
& 0.625$\pm$0.037
& 0.667$\pm$0.038
& 0.654$\pm$0.025
& 0.594$\pm$0.027
& 0.635 \\
OTSurv
& 0.637$\pm$0.028
& 0.638$\pm$0.049
& 0.621$\pm$0.012
& 0.643$\pm$0.025
& 0.635 \\
SurvAgent
& \textbf{0.683$\pm$0.022}
& 0.676$\pm$0.036
& \underline{0.695$\pm$0.013}
& \underline{0.676$\pm$0.052}
& \underline{0.683} \\
\midrule
\textbf{PathLoop}
& 0.671$\pm$0.037
& \underline{0.691$\pm$0.042}
& \textbf{0.735$\pm$0.005}
& \textbf{0.682$\pm$0.044}
& \textbf{0.695} \\
\bottomrule
\end{tabular}}
\end{table}

\begin{figure*}[!t]
\centering
\includegraphics[width=\textwidth]{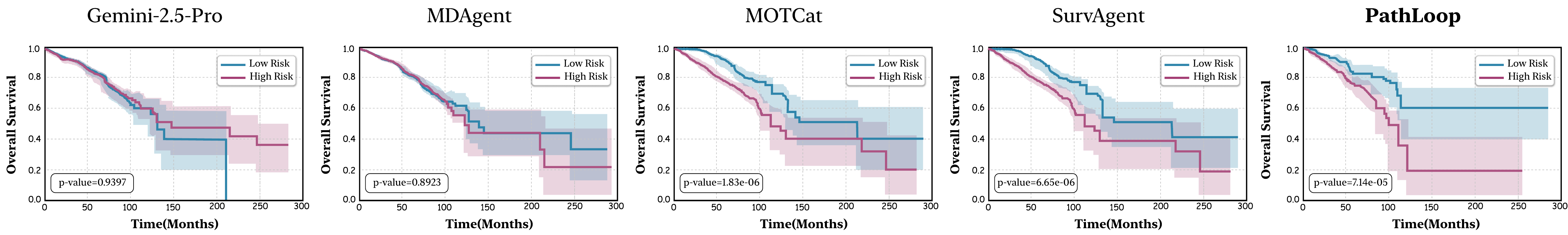}
\caption{Kaplan-Meier Analysis of predicted high-risk and low-risk groups. Shaded areas refer to the confidence intervals.}
\label{fig4}
\end{figure*}

\paragraph{Survival stratification (Figure~\ref{fig4}).}
Kaplan--Meier analysis shows that Gemini-2.5-Pro and MDAgent fail to produce a significant separation between high- and low-risk groups ($p>0.05$).
In contrast, PathLoop achieves a statistically significant stratification.
This indicates that our method captures stronger and more robust prognostic signals from WSIs and omic features, surpassing both general-purpose and agent-based alternatives in discriminative power.

\begin{figure*}[!t]
\centering
\includegraphics[width=\textwidth]{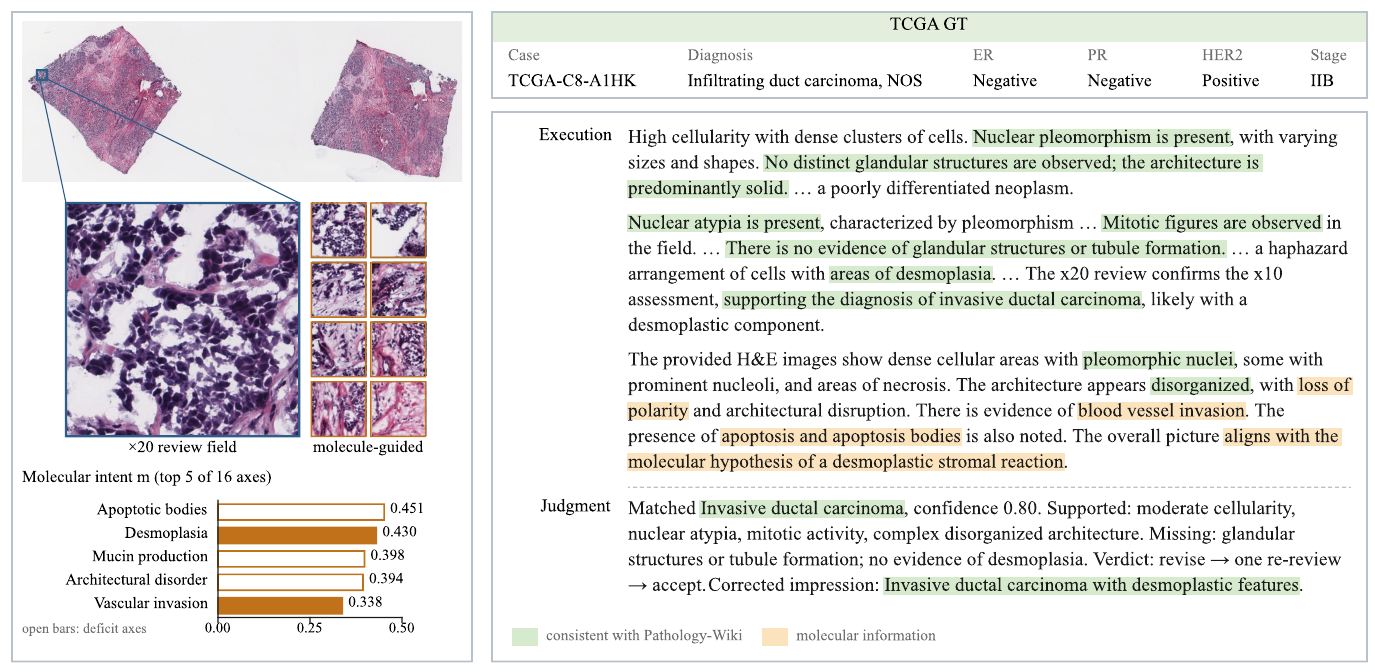}
\caption{Explainability analysis through visualization on case TCGA-C8-A1HK.}
\label{fig5}
\end{figure*}

\begin{table}[t]
\centering
\caption{Ablation study of the agent modules on TCGA-BRCA.
The full model achieves a C-index of 0.735.}
\label{tab:module_ablation}
\renewcommand{\arraystretch}{1.15}
\begin{tabular*}{\columnwidth}{
@{\extracolsep{\fill}}
llc
@{}
}
\toprule
Module & Setting & C-index \\
\midrule
Planning  & w/o CLS token      & 0.704 \\
Execution & w/o Projector      & 0.676 \\
Execution & w/o Omics          & 0.642 \\
Judgment  & w/o Pathology-Wiki & 0.648 \\
Persist   & w/o Memory         & 0.703 \\
\bottomrule
\end{tabular*}
\end{table}

\subsection{Ablation Study}
\paragraph{Module ablations (Table~\ref{tab:module_ablation}).}
Removing any single component degrades performance below the full model's C-index of 0.735, confirming that all four agents contribute to the final prediction.
The omics branch is the largest contributor: without it, the C-index falls to 0.642, underscoring the value of molecular priors for prognostic reasoning.
We attribute this drop to retrieval quality: without molecular priors, region selection reverts to visual saliency and misses inconspicuous but prognostically relevant areas, such as stromal reaction or subtle architectural atypia.
Judgment follows closely, with performance dropping to 0.648 when Pathology-Wiki verification is removed, indicating that knowledge-grounded correction recovers errors that visual evidence alone cannot.
Without it, overconfident draft diagnoses pass unchallenged, as exemplified by the case in Figure~\ref{fig5}.
Removing the projector yields 0.676, showing that slide-level context complements pixel-level reading.
Memory and the CLS token offer smaller but consistent gains, dropping to 0.703 and 0.704 when removed.

\paragraph{Impact of concept embedding $W$ and number of concepts $K$ (Table~\ref{tab:setting_ablation}).}
Our learned $W_{\mathrm{ours}}$ yields the best ER, PR, HER2, and Risk.
Corrupting $W$ with Gaussian noise ($W_{\mathrm{noise}}$) severely degrades HER2 and Risk, although ER and PR remain relatively robust.
Shuffling the pathway assignments ($W_{\mathrm{shuffle}}$) retains strong Subtype performance but reduces HER2 by $0.187$ and Risk by $0.051$, supporting that $W$ encodes structured biological knowledge rather than arbitrary connectivity.
Using a random $W$ degrades all endpoints, with a milder HER2 drop but still markedly below our learned representation.
Scaling the number of concepts from $K{=}8$ to $K{=}32$ brings consistent gains:
$K{=}32$ achieves the highest ER, HER2, and Risk, while $K{=}16$ reaches the best PR and Subtype.
$K{=}8$ severely underperforms, indicating that too few concepts merge clinically distinct patterns and hurt all tasks.

\begin{table}[t]
\centering
\renewcommand{\arraystretch}{1.1}
\caption{Ablation study of different settings on TCGA-BRCA.}
\label{tab:setting_ablation}

\begin{tabular*}{\columnwidth}{
@{\extracolsep{\fill}}
lccccc
@{}
}
\toprule
Setting & ER & PR & HER2 & Subtype & Risk \\
\midrule
$W_{\mathrm{ours}}$
& \textbf{0.964}
& \textbf{0.958}
& \textbf{0.854}
& \underline{0.983}
& \textbf{0.735} \\
$W_{\mathrm{noise}}$
& \underline{0.939}
& 0.897
& 0.578
& 0.975
& 0.685 \\
$W_{\mathrm{drop}}$
& 0.935
& 0.861
& 0.770
& 0.900
& \underline{0.725} \\
$W_{\mathrm{shuffle}}$
& 0.912
& \underline{0.897}
& 0.667
& \textbf{0.989}
& 0.684 \\
$W_{\mathrm{random}}$
& 0.898
& 0.764
& \underline{0.797}
& 0.893
& 0.686 \\
\midrule
$K=8$
& 0.577
& 0.745
& 0.523
& 0.659
& 0.563 \\
$K=16$
& \textbf{0.964}
& \underline{0.958}
& \textbf{0.854}
& \underline{0.983}
& \textbf{0.735} \\
$K=32$
& \underline{0.875}
& \textbf{0.962}
& \underline{0.673}
& \textbf{0.989}
& \underline{0.686} \\
\bottomrule
\end{tabular*}
\end{table}

In Table~\ref{tab:classifier_survival_sensitivity}, we examine the regularization strength $C$.
$C{=}0.1$ yields the best overall Risk and HER2;
increasing $C$ to $1.0$ improves Subtype (\textbf{0.992}) but reduces HER2 by $0.051$ and Risk by $0.037$, while PR is insensitive to $C$.
All reported results use $C{=}0.1$; the Cox model is kept with default L2 penalty, and predictions remain stable across the tested hyperparameter range.

\begin{table}[t]
\centering
\tiny
\caption{Sensitivity analysis of logistic regression.}
\label{tab:classifier_survival_sensitivity}
\setlength{\tabcolsep}{2.5pt}
\renewcommand{\arraystretch}{1.1}
\resizebox{\columnwidth}{!}{
\begin{tabular}{lcccccc}
\midrule
LogReg $C$
& ER
& PR
& HER2
& Subtype
& Risk \\
\midrule
0.1
& \textbf{0.964}
& \textbf{0.958}
& \textbf{0.854}
& \underline{0.983}
& \textbf{0.735} \\
0.5
& \underline{0.954}
& \textbf{0.958}
& \underline{0.804}
& 0.931
& 0.691 \\
1.0
& 0.881
& \textbf{0.958}
& 0.803
& \textbf{0.992}
& \underline{0.698} \\
\midrule
\end{tabular}}
\end{table}

\subsection{Visualization}
Figure~\ref{fig5} illustrates the closed-loop collaboration of planning, execution, judgment, and persistence agents.
The Planning Agent retrieves memories S3 (0.639) and D8 (0.549) and generates an 8-step plan.
The Execution Agent screens at $\times 2.5$, projects the 768-dim TITAN vector to a global CLS token, and narrows 63 suspicious patches to 8 (set $A$).
The Gene Interpreter encodes methylation/miRNA into a 16-dim morphological intent and retrieves set $B$; the intersection $A \cap B$ yields 8 patches.
Molecular priors assign high scores to apoptotic bodies, desmoplasia, mucin, architectural atypia, and vascular invasion, yet patch--gene cosine similarity is only $\cos(m,v)=0.2609$, indicating weak visual evidence.
The fused global token and $\times 20$ patches produce a draft diagnosis ``desmoplastic breast carcinoma''.
Visual consistency gating passes ($c=0.836$), but the Judgment Agent audits against a pathology knowledge base, finds insufficient support for tubule formation and desmoplasia, and issues \textsc{revise}.
After re-examination, consistency drops to $c=0.751$, while Judgment again demands revision.
Reconciling visual evidence with molecular priors, the system matches the case to invasive ductal carcinoma with confidence 0.80 and refines the diagnosis to ``invasive ductal carcinoma with desmoplastic features''.
The Persist Agent stores the corrected reasoning trace, skills, tools, and decisions into persistent memory.
This case demonstrates that visual self-consistency can retain a plausible but poorly supported diagnosis, whereas multi-agent evidence auditing identifies cross-modal inconsistency, corrects the diagnosis, and converts the correction into reusable long-term experience.

\section{Conclusion}
In this work, we proposed PathLoop, a self-evolving multi-agent framework that couples omics-guided region retrieval with cross-scale visual analysis, verifies diagnoses against Pathology-Wiki, and consolidates past cases into a four-level memory.Experiments establish state-of-the-art performance, offering an evidence-grounded route to continually improving whole-slide image analysis.

{
    \small
    \bibliographystyle{ieeenat_fullname_unsrt}
    \bibliography{main}
}

\end{document}